\documentclass{article}
\usepackage{ijcai19}

\usepackage{times}
\usepackage{soul}
\usepackage{url}
\usepackage[hidelinks]{hyperref}
\usepackage[utf8]{inputenc}
\usepackage[small]{caption}
\usepackage{graphicx}
\usepackage{amsmath}
\usepackage{booktabs}
\usepackage{tcolorbox}
\usepackage{latexsym} 
\usepackage{dingbat}
\usepackage{balance}

\usepackage{xcolor}
\def\shorten{\looseness=-1} 
\newcommand{\myNum}[1]{(\emph{#1})}
\newcommand{\nsstitle}[1]{\vspace{4pt}\noindent\textup{\textbf{#1}}}

\title{ChatGPT versus Traditional Question Answering for Knowledge Graphs: \\ Current Status and Future Directions Towards Knowledge Graph Chatbots}

\author{
Reham Omar$^1$\and
Omij Mangukiya$^1$\and
Panos Kalnis$^2$\And
Essam Mansour$^1$\\
\affiliations
$^1$Concordia University, Canada, 
$^2$KAUST, Saudi Arabia\\
\emails
\{Fname\}.\{Lname\}@concordia.ca,
panos.kalnis@kaust.edu.sa
}

\newcommand{\qald}{QALD-9}
\newcommand{\kgqan}{\textsc{KGQAn}}
\newcommand{\chatgpt}{ChatGPT}
\newcommand{\galactica}{Galactica}

\begin{document}
\pagenumbering{arabic}

\maketitle

\begin{abstract}

Conversational AI and Question-Answering systems (QASs) for knowledge graphs (KGs) are both emerging research areas: they empower users with natural language interfaces for extracting information easily and effectively. Conversational AI simulates conversations with humans; however, it is limited by the data captured in the training datasets. In contrast, QASs retrieve the most recent information from a KG by understanding and translating the natural language question  into a formal query, supported by the database engine.  

In this paper, we present a comprehensive study of the characteristics of the existing alternatives, towards combining both worlds into novel KG chatbots. Our framework compares two representative conversational models, {\chatgpt} and {\galactica}, against {\kgqan}, the current state-of-the-art QAS. We conduct a thorough evaluation using four real KGs across various application domains, to identify the current limitations of each category of systems. Based on our findings, we propose open research opportunities to empower QASs with chatbot capabilities for KGs. All benchmarks and all raw results are available\footnote{\url{https://gitfront.io/r/user-2881158/pCcUsY8d2145/lmkg/}} for further analysis. 
\end{abstract}

\section{Introduction}

Both conversational AI and question-answering systems (\emph{QAS}), aim to develop effective natural language interfaces for accessing useful information. Conversational AI models simulate conversations with humans to achieve a specific task, or participate in general discussions~\cite{CAIYan18} \cite{converastionAgent} \cite{ZhongLWM21}.
Their answers are formatted mainly as human-like text with explanations to build discussions. QASs focus mainly on retrieving information from KGs by translating a question into a structured query, that fetches the recent information relevant to the question from an existing KG~\cite{KGQan}. 
In QASs, an answer has a structured format, e.g., CSV- or Excel-like format, with no explanations. 
Combining the capabilities of conversational AI and question-answering systems for KGs, has the potential to lead the development of advanced chatbot systems.  
We envision KG chatbots that will enable companies to: \myNum{i} meet the diverse needs of end users; \myNum{ii} maximize the usability of their KGs; and \myNum{iii} maintain the KGs with dynamic context, based on the interactions with users.\shorten 

There is a sharp growth in the use of conversational AI models, such as {\chatgpt}~\footnote{\url{https://openai.com/blog/chatgpt/}}.
The model is trained on a large collection of datasets from diverse data sources, to generate human-like answers. 
Several KGs are captured from some of these datasets. This paper answers the question: \textit{how does a language model, such as {\chatgpt}, compare against traditional QASs for KGs?}
Our survey also identifies the main capabilities that should be combined into a chatbot for KGs. 
This new generation of KG chatbot systems will be able to benefit from the structured information in a KG, to provide more accurate answers based on a specific domain, such as academics, finance, or  health care.

Future KG chatbot systems will need a QAS mechanism to update the response generated by language models with the recent information retrieved from a given KG. Thus, the user will get responses reflecting the most recent information related to their domain-specific questions. The state-of-the-art QAS for KG is {\kgqan} ~\cite{KGQan}, a universal QAS that can answer questions from arbitrary KGs of various application domains. In {\kgqan}, question-answering is formalized as a three-fold problem: question understanding, linking, and filtering of the final answers. For question understanding, {\kgqan} utilizes Seq2Seq pre-trained language models~\cite{seq2seq} trained independently of KGs, such as BART \cite{BART} and GPT-3 \cite{gpt}. For linking and filtering, {\kgqan} uses a just-in-time approach based on word embedding models, such as FastText~\cite{fasttext}.

This paper proposes a comparative framework to systematically review the main capabilities of conversational AI language models versus KG QASs, and their styles in answering questions. We used our framework to comprehensively evaluate the state-of-the-art language models, such as {\chatgpt} and {\galactica} \footnote{\url{https://galactica.org}}, versus the state-of-the-art QAS. Our evaluation uses four real KGs from various application domains. We identify the current limitations of each category. We also propose open research challenges and opportunities for developing a KG chatbot that incorporates the merits of conversational AI models and QASs.\shorten


\section{Background: Language Models and QASs}
\label{sec:Background}


Language models, such as BERT \cite{bert}, T5~\cite{t5}, and GPT-3 \cite{gpt}, solve the question-answering task using text generation. Some models need to be re-trained for specific tasks. Given the input, the pre-trained model will directly produce the output. 
Other models have to be fine-tuned using prompting, e.g., the model gets a description of the required task and a few examples; then it can  directly be used for generating the required output. Based on the training datasets, language models can be classified into: \myNum{i} general models, such as PaLM~\cite{palm}, OPT~\cite{opt} and GPT-NeoX-20B~\cite{gptneox}; and \myNum{ii} specific domains models, such as   {\galactica}~\cite{GALACTICA}, SciBERT~\cite{scibert} and BioMegatron \cite{BioMegatron}.
   
More advanced conversational AI models and dialogue agents have benefited from the recent advancement in language models to create chatbots that can answer questions while engaging with users in conversations. Examples of these chatbots are {\chatgpt},  a recent chatbot introduced by OpenAI, and LaMDA \cite{lamda},  a family of transformer-based \cite{transformers} language models for dialogue applications. Moreover, Sparrow~\cite{sparrow} is a dialogue agent trained using reinforcement learning with human feedback, and webGPT~\cite{webgpt} is a fine-tuned GPT-3 model that answers questions by allowing the language model to search the web. Meena~\cite{meena} is a chatbot trained to respond in a human-like way by replying using sensible responses.\shorten

{\chatgpt} is a chatbot based on language models for answering questions, asking for clarification, creating dialogues with the user, and dealing with follow-up questions. It can apply reasoning to correct its answer based on users' feedback. Moreover, {\chatgpt} can write, debug, and optimize code.
It is trained using reinforcement learning with human feedback similar to the training process of InstructGPT \cite{instructgpt}, but using a combination of \myNum{i} a new dialogue dataset, and \myNum{ii} InstructGPT dataset after transforming its format to a dialogue format.\shorten

{\galactica} is a general-purpose scientific language model trained on a large corpus of scientific data for multiple tasks, such as predicting citations, reasoning, question answering and predicting protein annotations. It is based on transformer models~\cite{transformers} that consist of only a decoder with some modifications. 
It was trained on massive datasets from different open-access scientific sources, such as papers and filtered common crawl. Its training datasets also included some general knowledge, such as Wikipedia.\shorten 

Traditional QASs for KGs divide the question-answering process into three phases: question understanding, linking, and execution. 
In the question understanding step, the systems extract key entities and relations from the given question. 
The entities and relations are organized in an intermediate representation of the question, usually in the form of a graph.
In the linking step, QASs map the extracted key entities and relations to vertices and predicates from the target KG. The last step is organizing the retrieved vertices and predicates to create a SPARQL query, execute it and perform an optional filtration step, where the most relevant answers are returned to the user.\shorten

EDGQA~\cite{EDGQA} uses Stanford core NLP parser~\cite{CoreNLP} to predict the constituency parsing tree of the question. Then it applies human-curated heuristic rules to transform the tree into a root-acyclic graph, called an entity description graph (EDG). In the linking step, the nodes of the EDG are linked to vertices in the target KG using different linking methods, such as Falcon~\cite{falcon}. These methods depend on building indices in a pre-processing phase, then using them to retrieve the required output. EDGQA also performs another pre-processing phase to extract all entity types from the KG for filtration. Finally, the semantically equivalent SPARQL query to the question $Q$ is created using the linked vertices, predicates, and entity type to the different phrases in $Q$. 

{\kgqan}~\cite{KGQan} is the current state-of-the-art system for question answering on KGs.  
Question understanding is formalized as a triple-patterns generation model, trained using Seq2Seq pre-trained models, such as BART~\cite{BART} and GPT-3. 
The triple patterns are converted to a graph structure called Phrase Graph Pattern (PGP). 
{\kgqan} performs just-in-time linking based on built-in indices in the KG engines.
It also uses word embedding models, such as FastText~\cite{fasttext}, to assess the semantic affinity between different phrases in the question and vertices and predicates in the KG. 
{\kgqan} does not depend on any pre-processing index, or models trained on the KG, to perform the linking. Thus, {\kgqan} can answer questions on an arbitrary KG as an on-demand service. 
\section{Comparative Framework}
\label{sec:Framework}

This section introduces our comparative framework towards chatbots for KGs. It also defines a unified assessment strategy to compare conversational AI language models and QA systems, e.g., deciding the correctness of the generated answers.

\subsection{Criteria Towards KG Chatbots}
There is a need for a comparative framework that evaluates the differences between conversational AI language models and QASs for the question answering (\emph{QA}) task on KGs. 
We defined fundamental criteria, such as the answer's correctness, robustness, consistency, explainability and advanced understanding of different questions, plus more KG-oriented criteria, such as incorporating recent information and generality across domains.    

\nsstitle{Correctness:}
KGs are constructed by capturing information and semantics from different data sources for a specific domain. The QA task on KGs focuses more on extracting accurate answers that can be used as input data for subsequent tasks.
Thus, the correctness of the provided answers is one of the essential criteria in our framework. 
We measure the correctness of the answers by comparing them to those included in a benchmark’s ground truth. 
Our framework utilizes the micro-F1 score as the primary metric for correct answers. QASs, such as EDGQA and {\kgqan}, output the answers in a format that can be directly used by the evaluation script provided by {\qald}~\footnote{{https://github.com/ag-sc/QALD/blob/master/6/scripts/evaluation.rb}}. 
This script automatically calculates the micro-F1 score of the systems and can be directly employed for the other datasets. 

Language models, such as  {\chatgpt} and {\galactica}, generate answers formatted differently than the answers in the KG benchmarks, as discussed more in Section~\ref{sec:assessment}. Thus, calculating the micro-F1 score for these models is performed manually. For example, the process may include comparing the provided answer to the labels of the KG vertices of the corresponding answers in the ground truth. To ensure a fair comparison with QA systems, we use the following definitions of the metrics, that are found in the {\qald} script: 

    \begin{equation}
        C = S \cap G
    \end{equation}
    \begin{equation}
        \label{eq:precision}
        m_{prec} = \frac{\Sigma_{i=1}^N |C_{qi}|}{\Sigma_{i=1}^N |S_{qi}|}
    \end{equation}
    \begin{equation}
        m_{recall} = \frac{\Sigma_{i=1}^N |C_{qi}|}{\Sigma_{i=1}^N |G_{qi}|}
    \end{equation}
    \begin{equation}
        \label{eq:f1}
        microF1 = \frac{2 \cdot m_{prec} \cdot m_{recall}}{m_{prec} + m_{recall} }
    \end{equation}

\noindent where $S_q$ is the set of answers generated by the system for question $q$, $G_q$ is the set of ground truth answers provided by the benchmark for question $q$, $C_q$ is the set of correct answers generated by the system for question $q$, $N$ is the number of questions in the benchmark, $m_{prec}$ is the micro-precision, and $m_{recall}$ is the micro-recall.

\nsstitle{Determinism:} KGs contain structured information that is queried using query languages, such as SPARQL, or Cypher. 
Thus, answering questions against a specific KG is expected to be deterministic, that is, for a given question, the generated answer will always be the same, assuming the same context. 
In contrast, language models encapsulate information seen in the training datasets, and there is a degree of non-deterministic behaviour in the generated answers. 
We consider this factor by running each question three times and calculating  correctness based on the best answers received.\shorten

\nsstitle{Robustness} is the ability of KG chatbots to tolerate erroneous input, such as a question with spelling, or grammar mistakes; these are common due to the free-form input from humans. 
Natural language models are trained to generate the correct answers, despite the possible mistakes. 
In construct, in formal query languages, such as SPARQL, any simple mismatch in the vertices, or predicates will lead to wrong answers. Therefore, robustness is a challenging measure for QASs.
Our framework evaluates  robustness by injecting spelling, or grammar mistakes in a subset of the benchmarks' questions and evaluating the correctness of the answers. 

\nsstitle{Explainability:} We define explainability as the ability to provide explanations, or additional details related to the question, which can benefit the user. 
The explanation increases the confidence of the users about the answer, as it explains the reasoning or the process followed to provide the answer. 
Explainability is common in language models. QASs, on the other hand, do not provide explanations, as their answers are extracted using SPARQL queries.
In our framework, this factor is evaluated manually by observing the outputs of the language models for all benchmark questions. 
\shorten 

\nsstitle{Question understanding} is the ability to understand a given question, regardless of the correctness of the answer. 
This criterion is essential, as users ask questions of different degrees of complexity, such as aggregate, temporal, or multiple-intentions questions. 
In QASs, question understanding is a separate phase followed by linking to entities in the KG. To evaluate the question understanding ability of such systems, we observe the output of the first phase in the pipeline.
In contrast, language models encapsulate question understanding within the wider output generation process. Therefore, for a language model, we analyze the generated answer to decide whether the model understood the question.

\nsstitle{Incorporating recent information:}
KGs are frequently updated by adding, deleting, or changing facts. For example, Wikipedia is updated at a rate of two edits per second\footnote{\small \url{https://en.wikipedia.org/wiki/Wikipedia:Statistics}}, which results in hundreds of thousands of edits per day in the corresponding Wikidata KG. 
Users typically expect the most up-to-date answer.
This criterion depends on the adopted process used to train the models, or the type indices used by the QASs. We consider this criterion in the evaluation.

\nsstitle{Generality across different domains} is one of the desirable criteria. We define it as the ability to support the question answering task on KGs of various domains without re-training on the target KG. Real KGs are of large scale. Thus, the training process for a specific KG will consume large amounts of time and computing resources. To evaluate generality, we use four real KGs of two domains and benchmarks, including human-curated questions of different complexities and styles.

\subsection{Fairness and Manual Assessment}
 \label{sec:assessment}
For the question answering task, language models and QASs may use different versions of the same knowledge, such as unstructured text versus structured data (graph). 
It is challenging to evaluate these models and systems based on the same criteria, due to the lack of benchmarks and a unified method for calculating the correctness of answers. 
To overcome these challenges, we performed  manual evaluation while considering multiple factors to ensure the fairness of the assessment.

For correctness, we need to evaluate the outputs of the QASs and language models against the ground truth answers, which are either strings, or vertices from the KG. For QASs, this is straight-forward.  
In contrast, language models produce mainly human-like text. 
We compare the answer generated by language models to the label of the vertex from the KG. For example, for the question: \textit{Give me all movies with Tom Cruise}, the output of the language models and the label of the ground truth is the titles of the movies, so they can be easily compared. However, for the question: \textit{Who founded Intel}, language models produce the answer in the form of a paragraph mentioning the founders with some context related to the company. 
We manually search for the expected answer within that text and compare it to the ground truth. Each benchmark deals with a specific version of KGs. Hence, when evaluating language models, we only consider the answers in the period covered by the KG. For example, a particular benchmark uses a KG based on information until 2016 only. Some models, such as ChatGPT, encapsulate information until 2021. We exclude any answers not included in the KG, to assess the precision of the model fairly.   

\section{Experimental Evaluation}
\label{sec:eval}

\begin{table*}[t]
 \caption{Results of the four benchmarks. A QA system, such as KGQAn, is designed to provide full answers with high precision across KGs of different domains. Language models do not provide correct answers for domain-specific questions, where the training datasets did not cover.\shorten}
  \label{tab:results}
  \centering
  
\begin{tabular}{lcccccc|cccccc}
\hline
 & \multicolumn{3}{c}{\textbf{{\qald}}}&
 \multicolumn{3}{c}{\textbf{{YAGO}}}&
 \multicolumn{3}{c}{\textbf{{DBLP}}} &
 \multicolumn{3}{c}{\textbf{{MAG}}}\\
 
 \textbf{System}&
 {\textbf{P}} &{\textbf{R}} &{\textbf{F1}}&
 {\textbf{P}} &{\textbf{R}} &{\textbf{F1}}&
 {\textbf{P}} &{\textbf{R}} &{\textbf{F1}}&
 {\textbf{P}} &{\textbf{R}} &{\textbf{F1}}\\ 
 
 \hline
 {EDGQA} &
31.30 & \textbf{40.30} & 32.00&
41.90 & 40.80 & 41.40&
8.00 & 8.00 & 8.00 & 
 4.00 & 4.00 & 4.00 \\

 {\kgqan} &
 49.81  & 39.39 &  \textbf{43.99}&
 48.48  & \textbf{65.22} &  55.62&
 \textbf{57.86} & \textbf{52.02} & \textbf{54.78} &
 \textbf{55.41} & \textbf{45.61} & \textbf{50.04}
 \\

 {\galactica} &
   14.19 & 2.31 & 3.97&
  9.92 &  3.48 & 5.15 &
 1.50 & 0.42 & 0.66& 
  18.02& 0.02  & 0.04 \\
 
 {\chatgpt} &
55.85 & 10.61 & 17.83&
 76.10 & 36.30 & 49.15&
 48.28 & 0.74 & 1.46& 
 28.13 & 0.04 & 0.07 \\

 {\chatgpt-Excel} &
   52.08 & 13.65 & 21.63 &
  79.95 & 43.25 & 56.14&
 45.76 & 2.85 & 5.37& 
  32.67 &  0.08 & 0.16 \\

{\chatgpt-Follow up} &
  \textbf{60.94} & 14.44 & 23.34&
 \textbf{87.185} & 52.29 & \textbf{65.56}&
 51.52 & 3.59 & 6.72 & 
  33.14 & 0.06 & 0.11 \\

 \hline
\end{tabular}
\end{table*}

\subsection{Compared Models and Systems}

In our survey, we evaluate 
two traditional QASs for KG:  {\kgqan}
and EDGQA\footnote{https://github.com/HXX97/EDGQA/}, against 
two language models: {\galactica}\footnote{https://github.com/paperswithcode/galai} and {\chatgpt}. 
{\kgqan} and EDGQA  are the state-of-the-art systems for question answering on KGs. 
{\galactica} has five models  with a different number of parameters. We used the standard model with 6.7B parameters. 
{\chatgpt} is a recent and very popular language model. We considered three variations of {\chatgpt}: 
\myNum{i} \textit{Default}: we perform three runs per question using the “regenerate response” button; 
\myNum{ii} \textit{Follow-up}: we ask a follow-up question that requests the entire list of answers instead of the examples provided, i.e., we first ask the question and, after retrieving the answer, we input \textit{continue} to get a more detailed list; and 
{\myNum{iii}} \textit{Excel}: we ask {\chatgpt} to return its answers in a structured format, like an excel sheet. At the start of each thread we write a pre-defined message and then, for each question, we append to it the prefix \textit{List out all}. The message is as follows:
\begin{tcolorbox}
You only form the answers to my questions as an excel table.
Do not write explanations. I will ask you questions, fill out a sheet
with the answer, and decide on rows and columns based on
the answer. Give also index to each row incrementally in
the sheet.
\end{tcolorbox}

We refer to these three variations as {\chatgpt}, {\chatgpt}-Follow up and {\chatgpt}-Excel. In our experiments, answers are classified into three categories: \myNum{i} \emph{Correct}, where {\chatgpt} provides correct answers with respect to the golden truth; {\myNum{ii}} \emph{Wrong}, where {\chatgpt} provides answers that do not match the golden truth; and {\myNum{iii}} \emph{No answer}, where  {\chatgpt} is unable to produce any answer for the given question. The main reason for the failure is either {\chatgpt} was not trained on the given information,  or it does not understand the entity we are asking about. 


\subsection{Benchmarks}
Our evaluation utilizes four different benchmarks: two on general-fact KGs, namely  DBpedia~\cite{DBpedia} and YAGO\footnote{\url{https://yago-knowledge.org/downloads/yago-4}}~\cite{yago}; and another two benchmarks on KGs related to academia, namely DBLP\footnote{\url{https://dblp.org/rdf/release/dblp-2022-06-01.nt.gz}} and the Microsoft Academic Graph (\emph{MAG})\footnote{\url{https://zenodo.org/record/4617285\#.YrNszNLMJhH}}. The general-fact KGs have facts about entities representing places and persons. DBLP and MAG contain facts about scientific publications, citations, authors and institutions. Each benchmark includes a set of English questions against one of these four KGs. Each question is annotated with the equivalent SPARQL query and the answers to the question. These answers are retrieved by executing the given SPARQL query. The benchmarks consist of different categories of questions, such as boolean questions whose answer is $True$ or $False$, listing questions where the answer should contain an exhaustive list of the coresponding KG entities, and WH-questions (e.g., WH-at, WH-o, etc.) that ask about people, places, languages, etc.

We use {\qald}~\cite{qald9}, the most challenging and widely used benchmark to evaluate QASs.  {\qald} uses the DBpedia KG, and consists of $150$ questions for testing. The questions of {\qald} are human-generated. The other benchmarks, namely YAGO, DBLP, and MAG, are recently introduced by \cite{KGQan}. Each of these benchmarks contains $100$ questions for testing, which are also human-generated.
The questions against YAGO are similar to the ones of  {\qald}, that is,  questions about people and places. Both DBLP and MAG benchmarks have questions related to citations and authors. 

\subsection{Performance Evaluation}
We compare the two models and systems using the four benchmarks. Table ~\ref{tab:results} summarizes the precision, recall and micro F1 score for each competitor in each benchmark. {\kgqan} achieve comparable results on the general KGs ({\qald} and YAGO) and the academic KGs (DBLP and MAG). {\chatgpt} performs significantly better on the general KGs compared to its performance on the academic KGs. {\chatgpt} does not add any irrelevant answers. Thus, {\chatgpt} is consistently achieving better precision than {\kgqan} on {\qald} and YAGO.  However, {\chatgpt} struggles in recall as it does not, by default, fully answer questions with a long list of answers. {\kgqan} utilized models trained independently of the target KG. Thus, {\kgqan} achieves better precision and recall for academic graphs than EDGQA, which is trained mainly on {\qald}. {\qald} and YAGO use the general KGs. Hence, EDGQA achieves comparable results on both.\shorten

To benefit from the full potential of {\chatgpt}, we tested using the Excel and Follow-up variations of {\chatgpt} to retrieve more answers to questions with a long list of answers. {\chatgpt}-Excel achieved better results as we ask explicitly to \textit{List out all} answers. As shown in Table ~\ref{tab:results}, the {\chatgpt}- Excel improved upon the default {\chatgpt} for all benchmarks. It even surpassed {\kgqan} in YAGO benchmark. In general, {\chatgpt}-Follow-up outperformed {\chatgpt}-Excel, as we enforced further the model to provide the full answers. This shows that to get the full potential of {\chatgpt}, you need to engage in a conversation with it. Similar to {\chatgpt}, {\galactica} did not perform well on list questions, significantly affecting its recall and the overall F1 score. The performance of language models and systems like EDGQA dropped significantly in answering questions against unseen information.  


\begin{table}[t]
    \caption{Performance of {\chatgpt} across three runs on {\qald}, where $C$ means that the question was answered correctly, $W$ means the question was not answered correctly and $NA$ means that {\chatgpt} failed to return any useful answer.}
  \label{tab:qaldruns}
  \centering
  
\begin{tabular}{lccc}
\hline
\textbf{Run 1} & \textbf{Run 2} & \textbf{Run 3} & \textbf{Count} \\
\hline
C & C & C & 83 \\
C & C & W & 1 \\
C & W & C & 1 \\
C & W & W & 3 \\
NA & C & C & 20 \\
NA & C & W & 2 \\
NA & NA & C & 1 \\
NA & W & C & 2 \\
W & C & C & 1 \\
W & C & W & 2 \\
W & W & C & 1 \\
\hline
NA & NA & NA & 4 \\
NA & NA & W & 1 \\
NA & W & NA & 1 \\
NA & W & W & 12 \\
W & W & W & 15 \\
\hline
\end{tabular}
\end{table}

\begin{table}[t]
    \caption{Performance of {\chatgpt} across three runs on {MAG}, where $C$ means that the question was answered correctly, $W$ means the question was not answered correctly and $NA$ means that {\chatgpt} failed to return any useful answer.}
  \label{tab:magruns}
  \centering
  
\begin{tabular}{lccc}
\hline
\textbf{Run 1} & \textbf{Run 2} & \textbf{Run 3} & \textbf{Count} \\
\hline
C & C & C & 15 \\
C & W & C & 1 \\
NA & NA & C & 1 \\
W & C & C & 1\\
W & C & W & 1\\
W & W & C & 1\\
\hline
NA & NA & NA & 53 \\
NA & NA & W & 1 \\
NA & W & W & 2 \\
W & W & NA & 1 \\
W & W & W & 23 \\
\hline
\end{tabular}
\end{table}

\subsection{Determinism}
For the same question, {\kgqan} always produces the same SPARQL query and answers, regardless of the number of runs attempted. To evaluate {\chatgpt}, we run each question three times. In some questions, there is a slight change in the explanation of the answer, but the answer is the same. In other questions, the output changed utterly. In some cases, {\chatgpt} in the first run does not provide any answer but returns a correct answer in the second run. In other cases, It does not give any answer in the first run, but in the next run, it can produce an answer, but it is a wrong answer. Tables~\ref{tab:qaldruns} and~\ref{tab:magruns} show the outputs of {\chatgpt} in the three runs for both {\qald} and MAG benchmarks. For lack of space, we removed the results of the other two benchmarks, which are similar to Tables~\ref{tab:qaldruns} and~\ref{tab:magruns}. For {\qald}, {\chatgpt} is deterministic, producing the same answer 68\%. For YAGO, it is deterministic in 85\% of the cases. For DBLP and MAG, it is deterministic for 94\% and 91\% of the cases, respectively. Overall, {\chatgpt} achieves a high percentage of determinism across the benchmarks, while {\kgqan} is more deterministic than the language model.

\begin{table*}[t]
    \caption{Number of questions answered correctly per category, where SF is single fact questions, SFT are single facts questions with type, MF are multi-facts questions, and B are boolean questions. {\chatgpt} has better performance than {\kgqan} in understanding complex questions, such as SFT and MF. However, {\kgqan} performs better than {\chatgpt} across different domains, i.e., general and academic KGs.}
  \label{tab:questionscount}
 
  \centering
  
\begin{tabular}{lcccccccccccccccc}
\hline
\textbf{Systems} &
\multicolumn{4}{c}{\textbf{{\qald} (150 Q)}} &
\multicolumn{4}{c}{\textbf{YAGO (100 Q)}} &  \multicolumn{4}{c}{\textbf{DBLP (100 Q)} } &
\multicolumn{4}{c}{\textbf{MAG (100 Q)}}\\


\textbf{Category} & \textbf{SF} & \textbf{SFT} & \textbf{MF} & \textbf{B} &
\textbf{SF} & \textbf{SFT} & \textbf{MF} & \textbf{B} &
\textbf{SF} & \textbf{SFT} & \textbf{MF} & \textbf{B} &
\textbf{SF} & \textbf{SFT} & \textbf{MF} & \textbf{B} \\

\textbf{\# of ques.} & \textbf{81} & \textbf{28} & \textbf{37} & \textbf{4} &
\textbf{87} & \textbf{6} & \textbf{6} & \textbf{1} &
\textbf{85} & \textbf{11} & \textbf{4} & \textbf{0} &
\textbf{75} & \textbf{7} & \textbf{16} & \textbf{2} \\

\hline
{\kgqan} & 
46 & 7 & 9 & 0 &
\textbf{61} & \textbf{5} & 2 & 0 &
\textbf{49} & \textbf{4} & \textbf{1} & \_ &
\textbf{40} & \textbf{2} & \textbf{9} & \textbf{2} \\

{\chatgpt-Follow up} &
\textbf{68} & \textbf{19} & \textbf{27} & \textbf{3} &
52 & 4 & 2 & \textbf{1} &
11 & 1 & 1 & \_ &
16 & 0 & 4 & 0 \\
\hline
\end{tabular}
\end{table*}

\subsection{Chatbot-oriented Criteria}
\nsstitle{Robustness:} We chose five random questions that {\chatgpt} can answer. For each question, we modified it by introducing different spelling and grammatical mistakes. We produced five versions for each question and ran {\chatgpt} on them. It understood and solved all the modified questions except one. QASs, including {\kgqan} and EDGQA, cannot achieve robustness as spelling mistakes lead to a mismatch between extracted entities from a question and the corresponding vertices in a KG.  

\nsstitle{Explainability:} {\kgqan} provides the user with the answer to the question. There is no way for the user to understand the process that led to receiving this answer. In contrast, {\chatgpt} provides an answer with a reasonable explanation in a human-like text. Sometimes, extra helpful information related to the answer is provided to the user.

\nsstitle{Question Understanding} 
Our comprehensive evaluation included 450 questions against different domains and KGs. We evaluated the correctness by manually comparing the generated answers to the ground truth. Through this process, we recognized that {\chatgpt} has a perfect understanding of questions, even for those it fails to answer. We observed multiple answers to questions it was unable to answer, e.g. for question: \textit{How many papers did Du Linna publish?} that {\chatgpt} fail to respond, its reply mentions:\shorten
\begin{tcolorbox}
It is possible that you can find information about an individual's publications by searching online or by contacting their place of employment or academic institution.\shorten
\end{tcolorbox}
This answer is provided for many questions to help the user find the required information. We conclude that {\chatgpt} understands the question, but its training datasets do not include the necessary information. QASs, including {\kgqan} and EDGQA, lack these capabilities. 

We also analyzed both {\chatgpt}-Follow up and {\kgqan} in answering questions of different linguistic complexity. Table \ref{tab:questionscount} summarizes our results and shows the number of correctly answered questions for each category per benchmark. This subset covers all questions in the benchmarks, where $SF$ are single fact questions, $SFT$ are single fact questions with type, $MF$ are multi facts questions, and $B$ are boolean questions. For {\chatgpt}, we consider the question correctly answered if it produces the correct answers in any of the three runs. Although {\chatgpt} has good performance concerning the number of questions answered in the general knowledge benchmarks, this is not reflected in the F1 score. This is due to the low recall, as shown in Table~\ref{tab:results}. 

\subsection{KG-oriented Criteria}
Unlike {\chatgpt}, {\kgqan} understands most of the questions of different types across the different domains and maintains comparable performance in precision, recall and F1 score. This proves the generality of {\kgqan} across different domains and users, who may express questions of variant complexity. As shown in Table~\ref{tab:questionscount},  {\chatgpt} has a good performance on the general benchmarks, {\qald} and YAGO, solving more than 50\% of the questions. However, it cannot work as well in the academic KGs solving a maximum of 20\% of the benchmark correctly. We can conclude that the data sources that {\chatgpt} is trained on do not contain enough academic information from DBLP and MAG KGs. 

{\kgqan} can deal with new and updated KGs instantaneously. The user needs to provide {\kgqan} with the URL of the SPARQL endpoint of the new graph. The {\kgqan} models are independent of the KGs. {\kgqan} uses these models to convert a question into a SPARQL query, which will be executed against the KG and return an answer incorporating the recent information. {\chatgpt} is a trained model, which answers questions based on the seen data in training. If the data source is updated, {\chatgpt} will require further training to answer questions about the updated information. 

\begin{table}[t]
    \caption{Comparison between {\chatgpt} and {\kgqan} according to the metrics of the proposed comparative framework.
     KG Chatbots needs a combination of language models’ abilities, such as robustness and explainability, with QAS's abilities, such as generality across domains and providing recent information.}
    \vspace*{-1ex}
  \label{tab:framework}
  \centering
  
\begin{tabular}{lcc}
\hline
\textbf{Metric} & \textbf{{\chatgpt}} & \textbf{{\kgqan}} \\
\hline
Correctness & low & high\\
Robustness & Yes &  No\\
Determinism & high & high \\
Explainability & Yes & No \\
Question Understanding & advanced & Basic\\
Incorporate recent information & No & Yes \\
Generality / different domains & low & high \\
\hline
\end{tabular}
\vspace*{-2ex}
\end{table}



\subsection{Discussion}
We summarize our comprehensive evaluation in Table~\ref{tab:framework} for {\chatgpt} and {\kgqan} based on our comparative framework. Both can answer a high percentage of questions on the general knowledge graph. {\chatgpt} has a lower recall for questions having a long list of answers. {\kgqan} has good performance accross different KGs, while {\chatgpt} could not answer most of the questions against DBLP and MAG. Both have Deterministic behaviour in answering questions. {\chatgpt} is slightly less deterministic than {\kgqan}, which incorporates recent information in the KG without needing further training. {\chatgpt} is trained on information from different data sources. Thus, it does not generalize to unseen domain information. In contrast, {\kgqan} can only retrieve answers from the target KG. {\chatgpt} has an outstanding ability in terms of explainability and robustness. There is a room for improvement  for QASs, such as {\kgqan}, in terms of explainability and robustness and question understanding.



\section{KG Chatbots: Open Research Challenges}
\label{sec:Open}

Advanced KG Chatbots could be developed by incorporating capabilities of language models, such as {\chatgpt}, in question understanding, robustness, and explainability, with the outstanding capabilities of a QA system, such as {\kgqan}, in incorporating recent information and generality across different domains. This section discusses several research challenges in advancing QA systems with these capabilities.  

\subsection{Dialogue Support for KGs}

The traditional QAS pipeline consists of three phases: {\myNum{i}} Question Understanding, {\myNum{ii}} Linking, {\myNum{iii}} Execution and Filtration. We need to modify question understanding and linking steps to handle the dialogue. For question understanding, most systems generate an intermediate representation of the question, e.g., a graph structure. Systems must enrich and maintain this intermediate representation as the dialogue progresses. It will need to specify if the new question in the conversation refers to an existing component or if we need to add a new node or edge to be up to date with the status of the conversation. It will also need to decide whether an answer to a previous question needs to replace a node in the graph. In this case, systems need to specify a new node representing the answer the user wants from the KG. QASs will have to link the newly added nodes and edges to vertices and predicates from the target KG.

The existing benchmarks for the QA task on KGs are not designed to evaluate dialogues. Instead, these benchmarks contain individual questions. The initial step of creating a chatbot for KGs is engaging in a conversation with the user. To achieve this, we need a new generation of dialogue benchmarks. The answers throughout the dialogue should be from the KG. These benchmarks can contain different kinds of questions. For example, a dialogue can be about information related to the same entity or entity type.

\subsection{Advanced Features}
\nsstitle{Question understanding:}
{\chatgpt} can understand many types of questions better than {\kgqan} in terms of the linguistic complexity. Benchmarks questions are classified into different types, as proposed in~\cite{lcquad2}. {\kgqan} can understand and solve multiple types of questions, such as: {\myNum{i}} \textbf{single fact:} question about one single piece of information, {\myNum{ii}} \textbf{single fact with type}: question about a single fact, but the type of the answer is explicitly mentioned in the question, {\myNum{iii}} \textbf{multi facts:} question asking about the answers using more than one fact, and {\myNum{iv}} \textbf{boolean:} question giving a fact and asking whether it is true or false. \shorten

Other categories that {\chatgpt} can solve but  {\kgqan} can not include: {\myNum{i}} \textbf{two intentions:} answers to these questions are more than one fact, e.g. \textit{Who is the wife of Barack Obama and where did he get married?}, {\myNum{ii}} \textbf{count:} questions asking about the number of answers instead of asking about the answer themselves, e.g.\textit{What is the number of Siblings of Edward III of England ?}. {\myNum{iii}} \textbf{temporal aspects}: questions asking about facts in a specific time frame, e.g. \textit{With whom did Barack Obama get married in 1992?} 

To prove {\chatgpt} 's ability to understand different questions, we randomly selected a sample of 10 questions per category from LCQuAD-2.0 dataset~\cite{lcquad2}. For the temporal and two intention questions, {\chatgpt} managed to understand all of them and answered 90\% of the questions correctly. For count questions, {\chatgpt} did not perform well despite its ability to understand questions. It did not produce any answer for 50\% of the questions and managed only to solve correctly 10\% of the count questions. {\kgqan} needs to improve its Seq2Seq model based on the pre-trained language models to support these question types.  

\nsstitle{Explanability:} 
QASs, such as {\kgqan}, extract final answers using a structured query fetching information from a KG in a structured format. These answers are not attached with any explanations or references to relevant information. Conversational AI models, such as {\chatgpt}, provide the user with explanations and additional information about the answer. For QASs, explainability can be in the form of showing the steps followed to produce the answer, i.e. converting the question to the intermediate representation, linking them and getting the answer. This will give the user a better understanding and judgment of the final answer. 
For example, if the user asks about a specific city, but the systems deal with it as a country, it gives a wrong answer. Without the explanatory steps, the user would not know the reason for the error. But if the system shows the process, the user will clearly understand the reasoning behind the received answer.

\nsstitle{User Feedback:} 
There is a need to support different types of user feedback. In the previous example, the user can provide suggestions to improve the quality of the answers, e.g., by clarifying that the question is asking for a city, not a country. In this sense, explainability can lead to having good user feedback and getting the correct answer. Another possible type of user feedback can help maintain or update the information in the KG. In this case, the user may ask a question to extract some information from the KG but notice that this information is outdated. KG chatbot may grant some users privileges to update the KG. Another possibility is that the KG chatbot uses the information given by the user in the context of the current conversation only. After the conversation ends, the system discards the information. Moreover, the KG chatbot could maintain this information as temporary information that needs to be validated before updating the KG. 

\nsstitle{Identifying Unanswerable Questions:} QASs may produce an empty answer for some questions. The user can not know whether the KG does not contain facts related to the question or the system failed to get any answer due to an error in the pipeline. The current benchmarks do not include many unanswerable questions. The KG chatbot should be able to identify unanswerable questions. It may also need to suggest where to find answers as that shows the user the question is understood, but the KG has no related facts.

\section{Conclusion}
\label{sec:con}

This paper presents a technical survey and comparative framework to formally review the differences between conversational AI models, such as {\chatgpt}, and traditional question-answering systems (QASs) for knowledge graphs. Our survey conducted a comprehensive evaluation, including four real KGs of different application domains and 450 English questions of various linguistic complexity. Our framework defined seven metrics for quantitative assessment in comparing models, such as {\chatgpt} and QASs. Our framework identifies the main capabilities required for KG chatbots. Furthermore, our paper highlights open research challenges in advancing and expanding existing QASs with some conversational AI models' abilities, such as dialogue support, robustness, and explainability.



\balance
\bibliographystyle{named}
\bibliography{bibiography}

\end{document}